\documentclass[10pt,twocolumn,letterpaper]{article}

\usepackage[pagenumbers]{iccv} 

\definecolor{iccvblue}{rgb}{0.21,0.49,0.74}
\usepackage[pagebackref,breaklinks,colorlinks,allcolors=iccvblue]{hyperref}

\usepackage[table]{xcolor}
\usepackage{longtable}

\def\paperID{*****} 
\def\confName{ICCV}
\def\confYear{2025}

\title{Hierarchical Reasoning with Vision-Language Models for Incident Reports from Dashcam Videos}

\author{
    Shingo Yokoi\quad Kento Sasaki\quad Yu Yamaguchi\\
    Turing Inc. \\
    {\tt\small \{shingo.yokoi, kento.sasaki\}@turing-motors.com} \\
}

\begin{document}
\twocolumn[{
  \renewcommand\twocolumn[1][]{#1}
  \maketitle
  \begin{center}
    \includegraphics[width=0.99\linewidth]{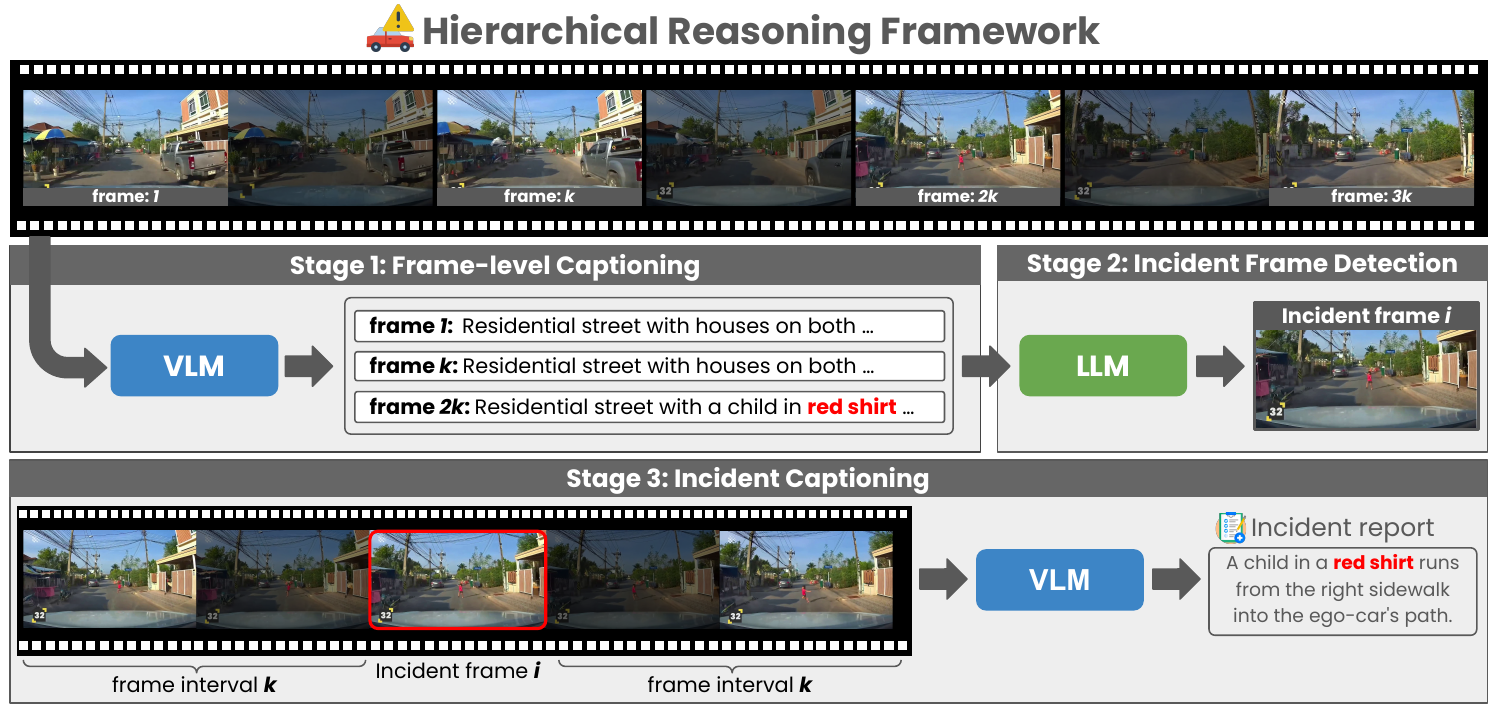}
    \captionof{figure}{Overview of our proposed hierarchical reasoning framework. The pipeline consists of three stages: (i) frame-level captioning, (ii) incident frame detection, and (iii) incident captioning, which together generate coherent incident reports from dashcam videos.}
    \label{fig:method}
  \end{center}
}]
\begin{abstract}

Recent advances in end-to-end (E2E) autonomous driving have been enabled by training on diverse large-scale driving datasets, yet autonomous driving models still struggle in out-of-distribution (OOD) scenarios. The COOOL benchmark targets this gap by encouraging hazard understanding beyond closed taxonomies, and the 2COOOL challenge extends it to generating human-interpretable incident reports. We present a hierarchical reasoning framework for incident report generation from dashcam videos that integrates frame-level captioning, incident frame detection, and fine-grained reasoning within vision-language models (VLMs). We further improve factual accuracy and readability through model ensembling and a Blind A/B Scoring selection protocol. On the official 2COOOL open leaderboard, our method ranks 2nd among 29 teams and achieves the best CIDEr-D score, producing accurate and coherent incident narratives. These results indicate that hierarchical reasoning with VLMs is a promising direction for accident analysis and for broader understanding of safety-critical traffic events. The implementation and code are available at \href{https://github.com/riron1206/kaggle-2COOOL-2nd-Place-Solution}{https://github.com/riron1206/kaggle-2COOOL-2nd-Place-Solution}.

\end{abstract}    
\section{Introduction}
\label{sec:introduction}

End-to-End (E2E) approaches have emerged as a prominent paradigm in autonomous driving~\cite{chen2024end}. These models are typically trained on large-scale multimodal datasets such as KITTI~\cite{kitti} and nuScenes~\cite{caesar2020nuscenes}, which are constructed from safe driving logs. In practice, real-world environments inevitably involve long-tail scenarios, encompassing rare events such as near-misses and other unpredictable incidents. Since such cases are largely absent from standard datasets, they constitute out-of-distribution (OOD) regimes for learned policies. Consequently, robustness to OOD hazards has been recognized as a critical requirement for the safe deployment of autonomous driving systems~\cite{yang2024generalized, bogdoll2022anomaly}.

To tackle this challenge, the COOOL benchmark~\cite{coool} was introduced to advance hazard understanding beyond closed taxonomies, fostering the detection, recognition, and prediction of both known and novel risks. Building upon this foundation, the 2COOOL challenge~\cite{2coool} further extends the task from hazard recognition to generating incident reports from dashcam videos, aiming to produce human-interpretable and consistent narratives of what occurred and why.

To this end, we propose a hierarchical reasoning framework with vision–language models (VLMs) for incident report generation. 
An initial attempt at naively processing entire videos with VLMs proved computationally expensive and frequently overlooked critical events, motivating a decomposition into smaller, more focused stages. 
Our proposed framework integrates frame-level captioning, incident frame detection, and fine-grained reasoning to produce accurate and coherent reports. 
These results highlight the potential of VLM-based hierarchical reasoning to advance reliable incident analysis and foster a broader understanding of safety-critical traffic scenarios.
\section{2COOOL Challenge}
\label{sec:challenge}

The 2COOOL Challenge~\cite{2coool} is part of the \textit{2nd Workshop on the Challenge of Out-of-Label Hazards in Autonomous Driving} (ICCV 2025) and aims to advance research on incident and hazard understanding from dashcam videos.

In the prior challenge, the COOOL Challenge~\cite{coool} introduced the concept of detecting and describing hazards through anomaly detection and open-set recognition. Building on this foundation, 2COOOL shifts the focus toward the automatic generation of incident reports. The dataset integrates three distinct resources, including COOOL~\cite{coool}, DADA~\cite{dada}, and Nexar~\cite{nexar}, which together cover a broad spectrum of driving scenarios, particularly unusual and safety-critical events. Each clip is recorded at 30 fps and lasts from a few seconds to several tens of seconds.

\subsection{Annotation Protocol}
To enable comprehensive incident understanding, the 2COOOL dataset provides contextual annotations for each dashcam clip. The annotation schema includes:
(i) event type (hazard, accident, or no incident);
(ii) crash severity;
(iii) ego-vehicle involvement;
(iv) counts of other involved entities (vehicles, pedestrians, cyclists or scooters, animals);
(v) time-to-hazard in frames or seconds; and
(vi) detailed captions describing the moments preceding and following the incident.
In addition, driver gaze information and gaze-based captions are incorporated. To ensure diversity and reliability, annotations were generated by VLMs and subsequently verified by human validators.

\subsection{Tasks}
To decompose the incident report generation problem, the 2COOOL Challenge defines a series of prerequisite tasks that provide the essential components for generating the final report:

\noindent\textbf{Time-to-Incident Start Estimation}:
Predict the frame or timestamp at which a situation becomes hazardous, thereby estimating the incident onset.

\noindent\textbf{Incident Detection}:
Classify each video as containing a hazard, an accident, or no incident.

\noindent\textbf{Incident Recognition}:
Determine the specific type of hazard or accident (e.g., jaywalking pedestrian, road debris, vehicle running a red light).

\noindent\textbf{Ego-Car and Other Parties Involvement}:
Identify whether the ego-vehicle is involved and specify the presence and counts of other participants (vehicles, pedestrians, cyclists or scooters, animals).

\noindent\textbf{Crash Severity}:
Assess the level of danger associated with the incident according to predefined severity levels.

\noindent\textbf{Caption Before the Incident}:
Provide a caption describing the video segment immediately preceding the incident.

\noindent\textbf{Caption After the Incident}:
Provide a caption explaining the cause or outcome of the accident.

By combining the outputs of these tasks, VLMs can generate detailed, context-rich incident reports. The ultimate goal is to produce coherent and human-interpretable narratives that not only describe what happened but also explain why it occurred.

\subsection{Evaluation Metrics}

The official leaderboard will report scores for each evaluation category described in the challenge. Final rankings will be determined by the average of CIDEr-D~\cite{vedantam2015cider}, METEOR~\cite{banerjee-lavie-2005-meteor}, and SPICE~\cite{anderson2016spice} computed on the submitted reports. In addition, a subset of finalist submissions will undergo blind review by organizers without conflicts of interest, who will assess both the ground-truth labels and the corresponding video footage. This dual evaluation protocol ensures that systems are judged not only by textual overlap with references but also by the factual accuracy, clarity, and practical usefulness of their incident descriptions.

\section{Method}
\label{sec:method}

In this section, we present our method for generating coherent incident reports from dashcam videos. Section~\ref{subsec:cfr} describes the hierarchical reasoning framework, Section~\ref{subsec:ensembling} outlines the ensembling strategy, and Section~\ref{subsec:abscoring} details the Blind A/B Scoring procedure.

\subsection{Hierarchical Reasoning Framework}
\label{subsec:cfr}

The Hierarchical Reasoning Framework, illustrated in Figure~\ref{fig:method} is composed of three modules, which are frame-level captioning, incident frame detection, and incident captioning. Through hierarchical analysis of dashcam videos, the framework identifies critical segments and supports the generation of interpretable and comprehensive incident reports.

\subsubsection{Stage 1: Frame-level Captioning}

The first stage is frame-level captioning. Incident or hazard videos typically last from a few seconds to several tens of seconds, and directly feeding the entire sequence into a VLM would result in prohibitive computational costs. To address this, we sample the video every $k$ frames and extract the last frame of each segment as a reference frame. Each reference frame is individually input to a VLM to generate a local caption representing its surrounding segment. To incorporate gaze information, each reference frame is augmented by vertically concatenating the raw video frame with its corresponding gaze heatmap. In addition to captions, the model also outputs metadata regarding incident-related objects such as pedestrians and animals. This approach reduces the number of visual tokens while preserving essential visual characteristics, thereby facilitating efficient caption generation and metadata extraction. 

\subsubsection{Stage 2: Incident Frame Detection}

The second stage is incident frame detection. The captions and incident-related metadata generated from reference frames in Stage 1 are structured and provided as input to a Large Language Model (LLM). Based on these inputs, the LLM predicts the incident frame $i$. Since reference frames around an incident often contain descriptions of hazardous factors, the model can efficiently approximate the temporal range of the incident. By narrowing candidate frames according to reference-frame captions, the approach achieves a balance between computational efficiency and detection accuracy. 

\subsubsection{Stage 3: Incident Captioning}

The third stage is incident captioning. Once the incident frame $i$ is identified in Stage 2, it serves as an anchor, and frames within a start and end offset $t$ around it are considered. We define the sampled frame set as
\[
\mathcal{F}(i, k, t) = \{\, i + mk \mid m \in \mathbb{Z},\ -t \leq m \leq t \,\},
\]
where $k$ is the frame interval and $t$ is the start and end offset relative to $i$. The frames in $\mathcal{F}(i, k, t)$ are then input to a VLM to generate the incident report.

\subsubsection{Implementation Details}

We summarize the models used in our experiments in Table~\ref{tab:models_per_stage}. All experiments are conducted on 8 NVIDIA H100/H200 GPUs, with each video processed in only a few minutes across all stages. 

To generate multiple candidate reports, we refine the prompt design and inference settings. Specifically, in Stage 1, we set the frame interval to $k=10$, generating captions every 10 frames. In Stage 3, the frame interval is set to $k \in \{2,6,11,12\}$ and the start/end offset to $t \in \{6,8,10\}$.

\begin{table}[h]
    \footnotesize
    \centering
    \begin{tabular}{ll}
    \hline
    \textbf{Stage} & \textbf{Models} \\
    \hline
    Stage 1 & GLM-4.5V~\cite{glm45v} \\
    Stage 2 & GPT-OSS-120B~\cite{agarwal2025gpt} \\
    Stage 3 & GLM-4.5V~\cite{glm45v}, Qwen3-VL-235B-A22B-Thinking~\cite{Qwen2VL, Qwen-VL} \\
    \hline
    \end{tabular}
    \caption{Models used at each stage of our experiments.}
    \label{tab:models_per_stage}
\end{table}

\subsection{Ensembling}
\label{subsec:ensembling}

While the hierarchical framework produces effective incident reports, inconsistencies and minor errors may still arise across different inference settings. To further improve the quality of the outputs, we employ an ensembling strategy. Specifically, we collect multiple candidate reports generated for the same test sample under different settings and input them into a LLM, which rewrites them into a final coherent report. This procedure leverages the complementary strengths of individual candidates, resulting in incident reports that are both more accurate and more fluent. 

We use Qwen3-Next-80B-A3B-Instruct~\cite{qwen3technicalreport} for ensembling to consolidate multiple candidate reports into a single coherent output.

\subsection{Blind A/B Scoring}
\label{subsec:abscoring}

While automatic evaluation metrics provide a useful approximation of report quality, their outcomes do not always align with human judgment~\cite{sai2022survey, bertscore}. To more reliably identify methods that produce higher-quality incident reports, we employ Blind A/B Scoring. In this protocol, incident reports generated under different methods or settings are paired and presented in random order, with their origin concealed. For each pair, evaluators indicate their preference by selecting A, B, or Tie. Assessments are based on factual correctness, readability, and trustworthiness, which are jointly considered in the overall judgment. Each pair is evaluated by multiple annotators, and the final outcome is determined by majority vote. This process enables a robust comparison of methods and allows us to determine which approach produces more useful reports.

Figure~\ref{fig:abscoring} illustrates the interface of our web application for A/B Scoring.
\section{Results}
\label{sec:results}

The results of our candidate reports are summarized in Table~\ref{tab:internal_final_report}. Overall, the SPICE, METEOR, and CIDEr-D scores were generally aligned with human judgments, although notable differences remained. These findings highlight the importance of combining multiple quantitative metrics with human evaluation, as the final rankings in the 2COOOL Challenge are ultimately determined by organizers based on human assessment. For our submission, we selected the report that achieved the highest rating in blind A/B scoring as the final submission.

\begin{table}[h]
    \footnotesize
    \centering
    \begin{tabular}{l cccc c}
    \toprule
    ID & SPICE & METEOR & CIDEr-D & Final Score & A/B Ranking \\
    \midrule
    I   & 0.1717 & 0.2489 & 0.0054 & 0.1420 & 2 \\
    II & \underline{0.1739} & \underline{0.2547} & \underline{0.0063} & \underline{0.1449} & 1\textasteriskcentered \\
    \rowcolor{gray!30}
    III  & \textbf{0.1822} & \textbf{0.2605} & \textbf{0.0067} & \textbf{0.1498} & 1\textasteriskcentered \\
    \bottomrule
    \end{tabular}
    \caption{Comparison of our candidate reports. Final scores are computed as the average of SPICE, METEOR, and CIDEr-D. The A/B Ranking is derived from the results of Blind A/B Scoring by three human evaluators. * indicates no significant difference.}
    \label{tab:internal_final_report}
\end{table}

\begin{figure}[h]
    \centering
    \includegraphics[width=0.99\linewidth]{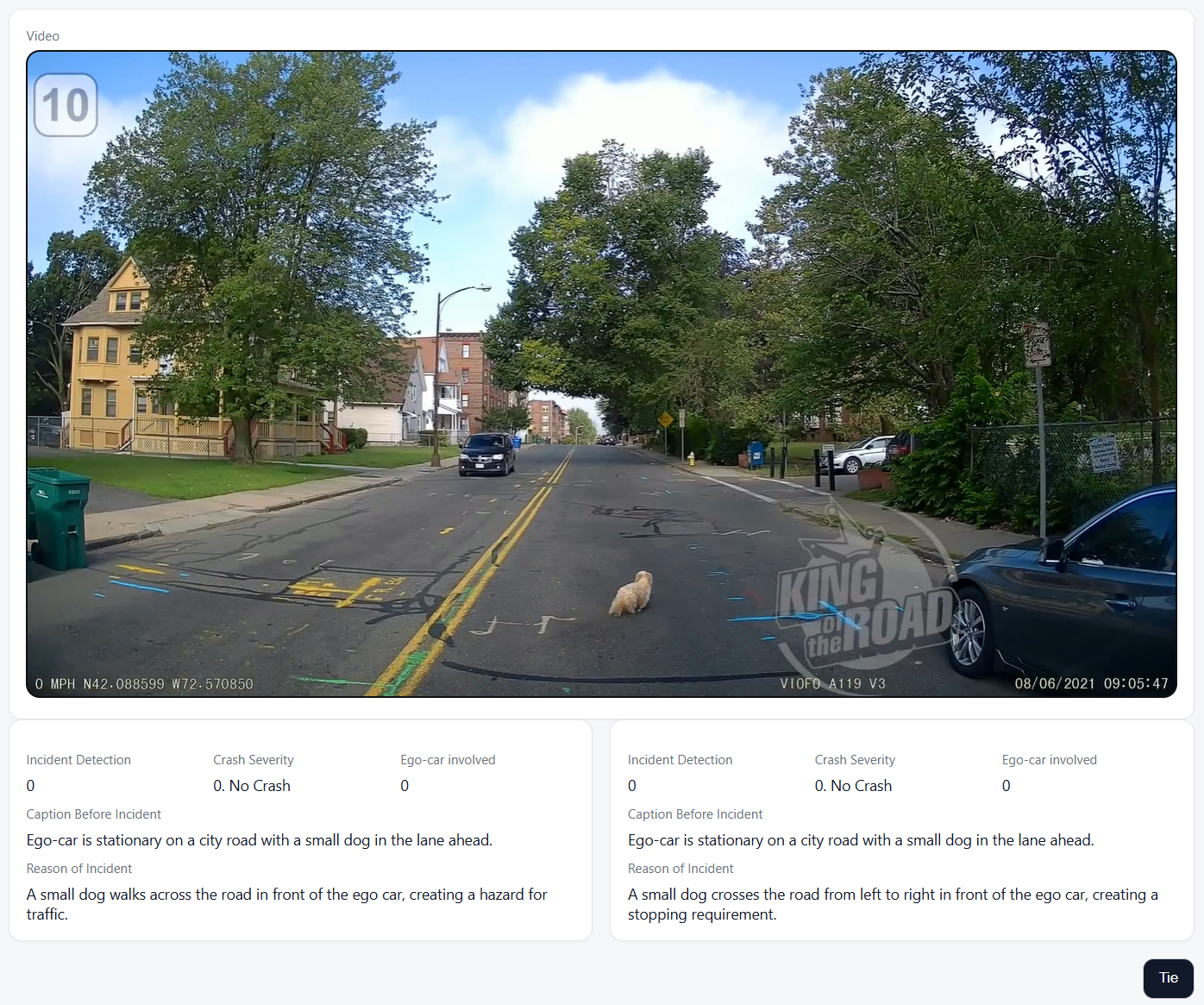}
    \caption{Interface of the web application for blind A/B scoring, where two reports are shown in random order and human evaluators select the more accurate one.}
    \label{fig:abscoring}
\end{figure}

An example of Blind A/B Scoring is shown in Figure~\ref{fig:abscoring}. In this case, the left-hand report stated that \textit{a small dog walks across the road}, whereas the right-hand report described the scene more precisely as \textit{a small dog crosses the road from left to right in front of the ego car}. The evaluator therefore selected the right-hand report.

\begin{figure}[h]
    \centering
    \includegraphics[width=0.99\linewidth]{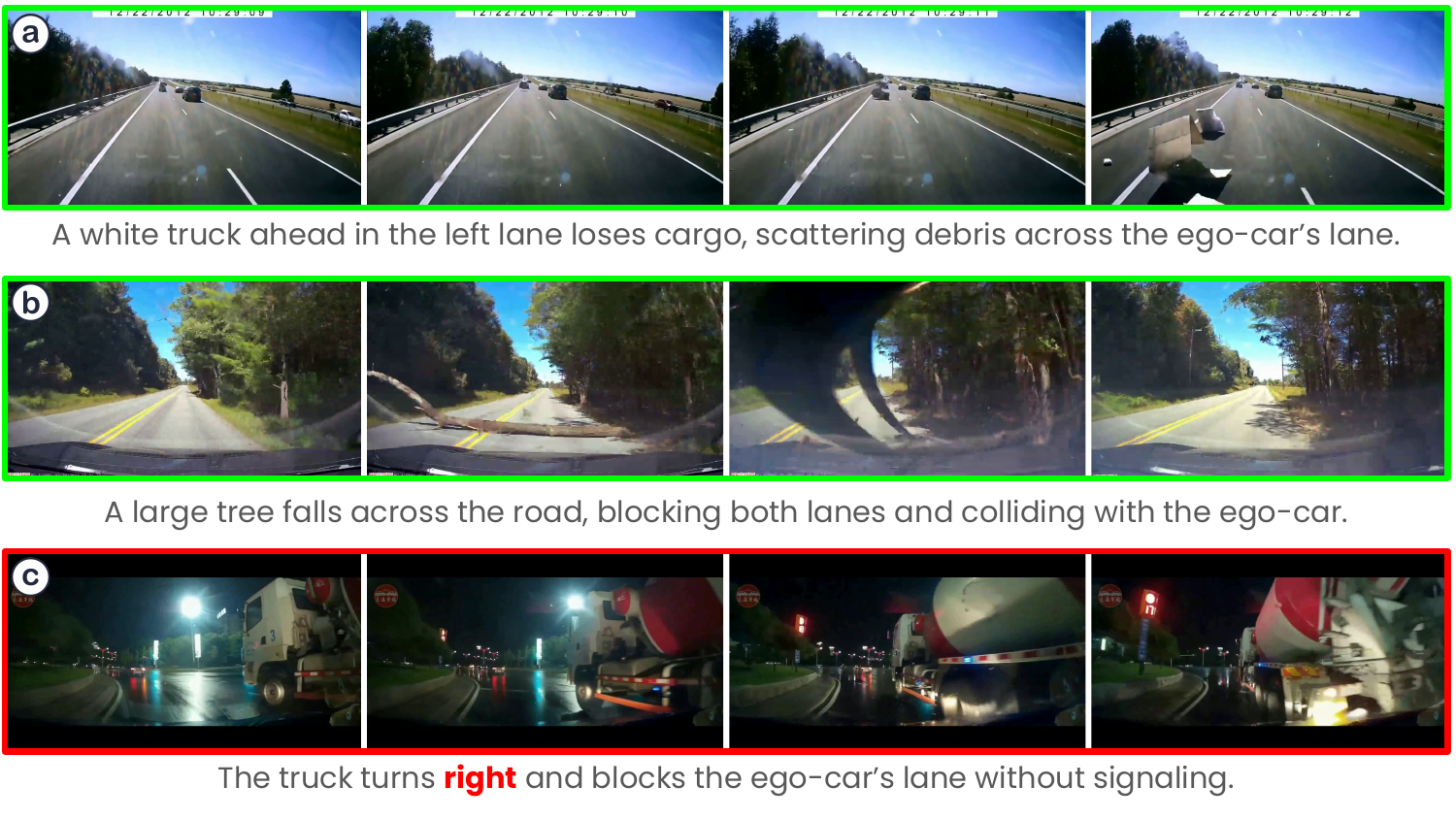}
    \caption{Qualitative examples of incident reports generated by our method. Most reports are factually accurate and coherent, such as (a) and (b), but occasional errors occur in relative spatial orientation, such as confusing right with left (c).}
    \label{fig:results}
\end{figure}

Figure~\ref{fig:results} presents qualitative examples of incident reports generated by our method. As illustrated in cases (a) and (b), our approach successfully produces accurate and coherent reports across a wide range of scenarios. However, as shown in case (c), errors occasionally arise in relative spatial expressions such as distinguishing left from right. These mistakes reflect a well-known limitation of current VLMs in spatial reasoning~\cite{strideqa, fu2024blink}, which remains an open challenge for future research.

Finally, Table~\ref{tab:final_report} shows the final scores of the top-ranked entries on the 2COOOL leaderboard at the end of the competition\footnote{\url{https://2coool.net}}. The scores among the top teams were very close, with our method ranking 1st on CIDEr-D and 2nd on the final score out of 29 entries on the open leaderboard.

\begin{table}[h]
    \footnotesize
    \centering
    \begin{tabular}{c l cccc}
    \toprule
    \# & Team Name & SPICE & METEOR & CIDEr-D & Final Score \\
    \midrule
    1 & NotSoDeep & \textbf{0.1911} & 0.2602 & 0.0040 & \textbf{0.1518} \\
    \rowcolor{gray!30}
    2 & Turing Inc. & 0.1822 & \underline{0.2605} & \textbf{0.0067} & \underline{0.1498} \\
    3 & Awais & \underline{0.1832} & \textbf{0.2614} & \underline{0.0046} & 0.1497 \\
    4 & Jane Doe & 0.1635 & \textbf{0.2614} & 0.0036 & 0.1428 \\
    5 & iAmAbIrD & 0.1596 & 0.2508 & 0.0028 & 0.1378 \\
    \bottomrule
    \end{tabular}
    \caption{Final scores of the top-ranked entries on the 2COOOL open leaderboard, computed as the average of SPICE, METEOR, and CIDEr-D.}
    \label{tab:final_report}
\end{table}

\section{Conclusion}
\label{sec:conclusion}

In this report, we introduced a hierarchical reasoning framework for generating incident reports from dashcam videos. The method integrates frame-level captioning, incident frame detection, and fine-grained reasoning to produce accurate and coherent reports. We further showed that ensembling and Blind A/B Scoring provide a principled selection mechanism for choosing the most accurate method. On the official open leaderboard of the 2nd Workshop on the Challenge Of Out-Of-Label Hazards in Autonomous Driving at ICCV 2025, our approach ranks 2nd out of 29 teams and achieves the best CIDEr-D score. Overall, our contribution demonstrates the potential of VLM-based hierarchical reasoning to advance reliable incident analysis and foster a broader understanding of safety-critical traffic scenarios.
{
    \small
    \bibliographystyle{ieeenat_fullname}
    \bibliography{main}
}

\end{document}